# TTNet: Multi-Task Deep Learning for Table Tennis Player Analysis with Smart Racket

Ko-Hsun Chen, Xiang-Wei Ke, Hsien-Cheng Huang, and Shang-Kuan Chen

Department of Computer Science and Engineering,
Yuan Ze University, Taoyuan, Taiwan
{s1111408,s1101327,s1103339}@mail.yzu.edu.tw
cotachen@saturn.yzu.edu.tw

**Abstract.** The AI CUP 2025 Precise Analysis of Table Tennis Smart Racket Data Competition has developed smart table tennis rackets that collect extensive player swing data, initiating research on table tennis big data. This allows for in-depth analysis of players' return methods and the consistency of swing force, improving the accuracy of player skill assessment. In response, this study focuses on six-axis sensor data from the smart table tennis racket and proposes a novel deep learning model with multitask learning capabilities, TTNet, to promote research and applications related to table tennis big data and tactical analysis. The model combines Convolutional Neural Networks (CNN), Residual Networks (ResNet), and Self-Attention mechanisms, and can simultaneously predict four tasks: gender, playing hand, years of experience, and player level. We implement a two-stage training strategy, incorporating data augmentation techniques and loss function design based on task characteristics, effectively enhancing the model's generalization ability on imbalanced data. Our approach ultimately achieved second place on the official leaderboard. The code is available here: https://github.com/ckexun/TTNet.git

## 1 Introduction

With the rapid development of sports technology, data-driven training methods have gradually become mainstream [1]. Sports such as basketball and football have widely adopted smart sensing and data analysis technologies as a basis for technical evaluation and training assistance. Table tennis, as a sport beloved by over one billion people worldwide, has more than two million participants in Taiwan and demonstrates outstanding performance in international competitions, leading to an increasing demand for precision training and analysis.

The AI CUP 2025 Precise Analysis of Table Tennis Smart Racket Data Competition has recently developed smart racket devices that enable real-time collection of high-frequency sensor data during player swings, including six-axis time series signals such as acceleration and angular velocity [2]. These data exhibit strong temporal characteristics and phase-specific variations, containing rich technical and habitual information. However, traditional statistical analysis or manual feature extraction methods struggle to fully capture the local details and global dependencies within the data, resulting in limited performance in identifying player attributes, styles, and player level.

This study aims to integrate sports technology and artificial intelligence by designing a deep learning framework specifically designed for smart table tennis racket sensor data. The model is capable of processing high-frequency and long time series sensor data while simultaneously predicting multiple key player attributes, including gender, playing hand, years of experience, and player level. This enables more comprehensive training assistance and technical evaluation, thereby facilitating the technological and precision-oriented development of table tennis, enhancing the practical utility of smart sports devices in real-world training applications.

## 2 Related Work

With the rapid development of sports technology and wearable devices, an increasing number of studies have applied deep learning techniques to motion recognition and sports analysis [3]. In table tennis research and related fields, high-frequency time series data captured by sensors provides rich motion signal features, serving as a crucial basis for identifying player movements, techniques, and habits.

In the context of action recognition and time-series data modeling, multitask learning has been demonstrated to effectively improve the overall performance and generalization ability of multi-label prediction in deep neural network architectures [4]. Moreover, the Transformer architecture, which is composed of self-attention mechanisms, has been widely adopted in sequence data analysis research due to its excellent capabilities for modeling long-range dependencies [5]. To enhance model robustness, studies have also proposed data augmentation techniques such as Mixup [6] and loss function designs like Label Smoothing [7] to improve model performance under conditions of unclear class boundaries and imbalanced data.

In the domain of smart table tennis applications, previous studies have combined wearable sensors with deep learning models, utilizing a modified LSTM architecture to predict technical scores for forehand strokes. This demonstrates the feasibility of using deep learning to analyze individual movement performance in table tennis [8]. However, this study focused only on score regression prediction for a single action, involved data from only 16 players and three stroke types, resulting in a relatively limited scope of application. The recently published TTSwing dataset further expands both the volume of data and the diversity of annotations, serving as a large-scale swing sensing dataset [9] that collects nine-axis sensing data and personal attribute labels from 93 players. It includes manually segmented sequences and engineered features, which are used for swing analysis and model training. In contrast, this study utilized large-scale raw six-axis sensor data and designed an end-to-end multitask deep learning model that can directly input continuous sensor waveforms, eliminating the need for additional segmentation and manual feature design, thereby demonstrating greater flexibility and deployment potential. Additionally, it outperforms previous studies on single-task action classification in terms of model generalization ability and application.

# 3 Methodology

In this section, we introduce our methodology, a multitask deep learning approach specifically designed to process high-frequency six-axis sensor data. The objective of this method is to automatically extract temporal features from table tennis swing actions through an end-to-end model structure, while simultaneously predicting multiple key player attributes.

## 3.1 Dataset

This study utilizes the AI CUP 2025 Precise Analysis of Table Tennis Smart Racket Data Competition Dataset [2]. The data is divided into two parts: a training set and a testing set, which includes public and private subsets. The training set contains 1,955 continuous swing test samples, and the testing set contains 1,430.

The training data consists of the train_info.csv label file and the corresponding train_data/ folder. The label file contains the unique_id, player_id, swing mode, gender, playing hand, years of experience, player level, and cut points for each swing, where:

- gender is labeled as 1 (male) or 2 (female).
- playing hand is 1 (right-handed) or 2 (left-handed).
- years of experience is 0 (low), 1 (medium), or 2 (high).
- player level is 2 (college division A player), 3 (college division B player), 4 (junior national team player), or 5 (junior player).

Each unique_id in the label file corresponds to a .txt file, and each .txt file contains the sensor data from a single player's swing test, including 27 consecutive swing records with pre- and post-swing disturbance. Each data point is a six-axis sensor time series consisting of three-axis acceleration (Ax, Ay, Az) and three-axis angular velocity (Gx, Gy, Gz). The data is a high-frequency continuous time series (see Fig. 2).

| unique_id | player_id | mode | gender | hold racket | play years | level | cut_point |
|---|---|---|---|---|---|---|---|
| 1 | 41 | 1 | 1 | 1 | 1 | 5 | [ 0 61 122 183 244 305 366 428 489 550 611 672 733 794 856 917 978 1039 1100 1161 1222 1284 1345 1406 1467 1528 1589 1651] |
| 2 | 41 | 2 | 1 | 1 | 1 | 5 | [ 0 74 149 224 299 374 449 524 599 674 749 824 899 974 1048 1123 1198 1273 1348 1423 1498 1573 1648 1723 1798 1873 1948 2023] |
| 3 | 41 | 3 | 1 | 1 | 1 | 5 | [ 0 103 207 311 415 519 623 727 831 935 1039 1143 1247 1351 1454 1558 1662 1766 1870 1974 2078 2182 2286 2390 2494 2598 2702 2806] |
| 4 | 41 | 4 | 1 | 1 | 1 | 5 | [ 0 101 203 304 406 507 609 710 812 913 1015 1116 1218 1319 1421 1522 1624 1725 1827 1928 2030 2131 2233 2334 2436 2537 2639 2741] |
| 5 | 41 | 5 | 1 | 1 | 1 | 5 | [ 0 105 211 317 423 529 635 740 846 952 1058 1164 1270 1376 1481 1587 1693 1799 1905 2011 2117 2222 2328 2434 2540 2646 2752 2858] |
| 6 | 41 | 6 | 1 | 1 | 1 | 5 | [ 0 104 208 312 416 521 625 729 833 938 1042 1146 1250 1354 1459 1563 1667 1771 1876 1980 2084 2188 2292 2397 2501 2605 2709 2814] |
| 7 | 41 | 7 | 1 | 1 | 1 | 5 | [ 0 111 222 334 445 556 668 779 890 1002 1113 1224 1336 1447 1558 1670 1781 1892 2004 2115 2226 2338 2449 2560 2672 2783 2894 3006] |
| 8 | 41 | 8 | 1 | 1 | 1 | 5 | [ 0 105 210 315 421 526 631 737 842 947 1052 1158 1263 1368 1474 1579 1684 1790 1895 2000 2105 2211 2316 2421 2527 2632 2737 2843] |
| 9 | 41 | 9 | 1 | 1 | 1 | 5 | [ 0 104 209 314 419 524 628 733 838 943 1048 1152 1257 1362 1467 1572 1677 1781 1886 1991 2096 2201 2305 2410 2515 2620 2725 2830] |
| 10 | 41 | 9 | 1 | 1 | 1 | 5 | [ 0 99 199 299 399 498 598 698 798 898 997 1097 1197 1297 1396 1496 1596 1696 1796 1895 1995 2095 2195 2294 2394 2494 2594 2694] |

**Fig. 1.** The structure of train_info.

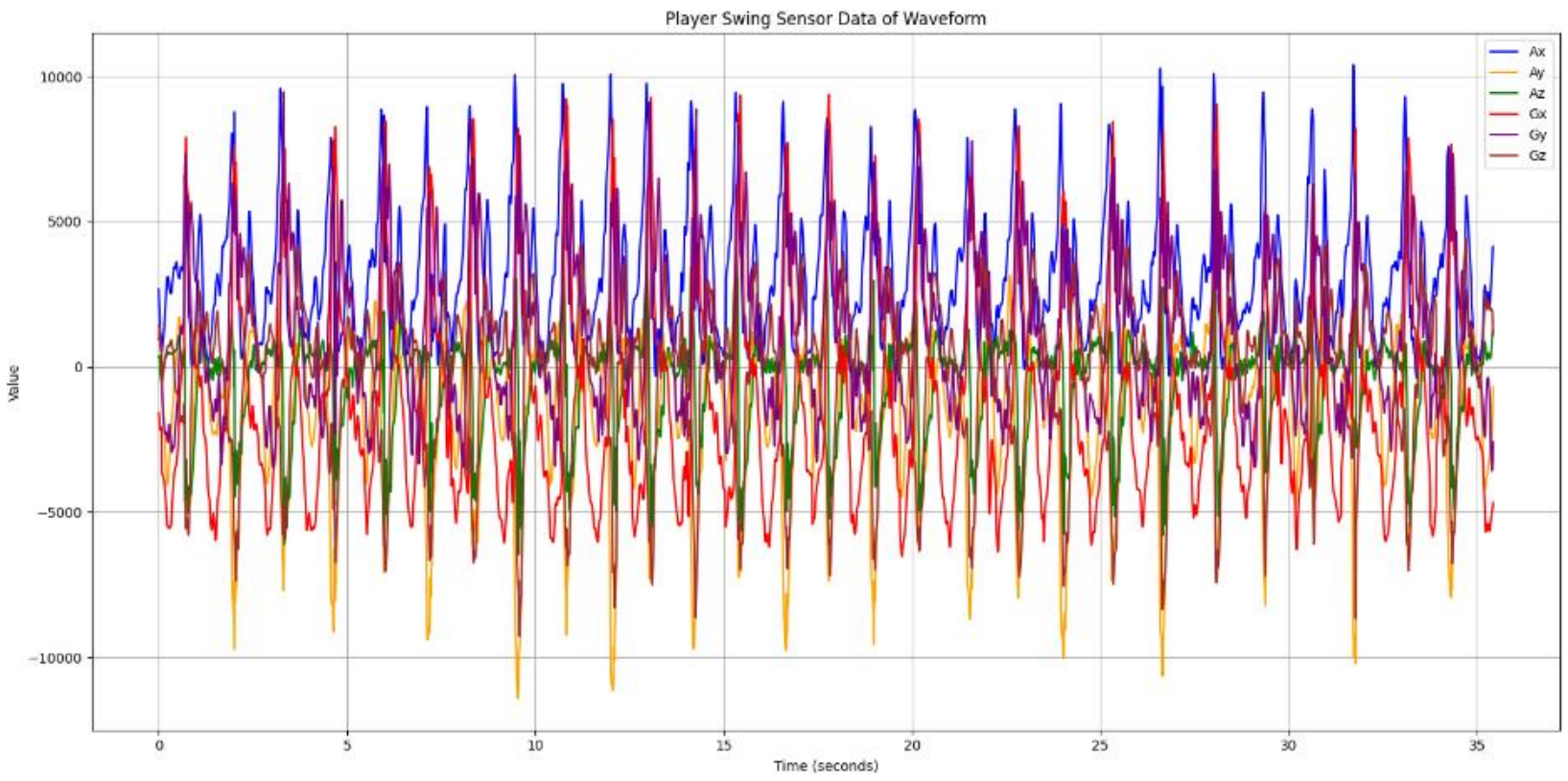

**Fig. 2.** Visualization of test record data.

### 3.2 Data Preprocessing

Based on the characteristics of the swing test data and to enhance the consistency and stability of model training, we applied several preprocessing steps to the input data, including label encoding, standardization, sequence alignment, and real-time data augmentation.

**Label Encoding.** In the original data, to align the model output format and avoid incorrect mapping, we uniformly convert the gender, playing hand, and player level label values to zero based encoding to correspond to the output range of sigmoid or softmax functions.

**Standardization and Sequence Alignment.** Each swing data is a high-frequency six-axis sensor sequence. Before feature learning, the data must be standardized and unified in length. We first apply Z-score standardization to process each data point, helping the model converge more stably during the training phase [10]. Next, since the length of the raw data for each player's swing varied, we preprocessed all the data and align it to a length of 2500. This length has been experimentally verified to achieve good results in model performance:

- If the sequence length is greater than 2500, randomly trim a segment of length 2500.
- If the sequence is less than 2500, apply zero padding at the end to make up the length.

**Data Augmentation.** To effectively improve the model's robustness and generalization ability, we introduced real-time data augmentation technology during training process[11, 12].

- Gaussian Noise: Add tiny Gaussian noise to the data with a 50% probability to simulate the small errors that may occur in sensor data and enhance the model's tolerance to environmental noise.
- Time Reverse: With a certain probability (30% for Stage 1 and 10% for Stage 2), the entire time series is reversed. This enhancement method assumes that many movement patterns identifying personal characteristics are reversible, enabling the model to learn the underlying characteristics independent of time direction.

– Mixup: With a defined probability, two samples and their corresponding labels are randomly selected and linearly interpolated to create synthetic training data. In our multitask setting, appropriate Mixup processing was applied to avoid label misalignment across different tasks.

### 3.3 Model Architecture

This study proposed a deep neural network model called TTNet, which is designed using multi-task learning [4]. The aim is to predict four different targets from the time series sensor data of table tennis players: gender, playing hand, years of experience, and player level.

Our model architecture combines CNN, residual connections, and self-attention mechanisms [5, 13, 14] to efficiently extract features from time-series data. The input consists of fixed-length six-axis time-series data, which are processed through multiple convolutional layers, residual blocks, and self-attention modules. The resulting representations are then used to perform classification across four player attributes. An overview of the model structure is illustrated in Fig. 3.

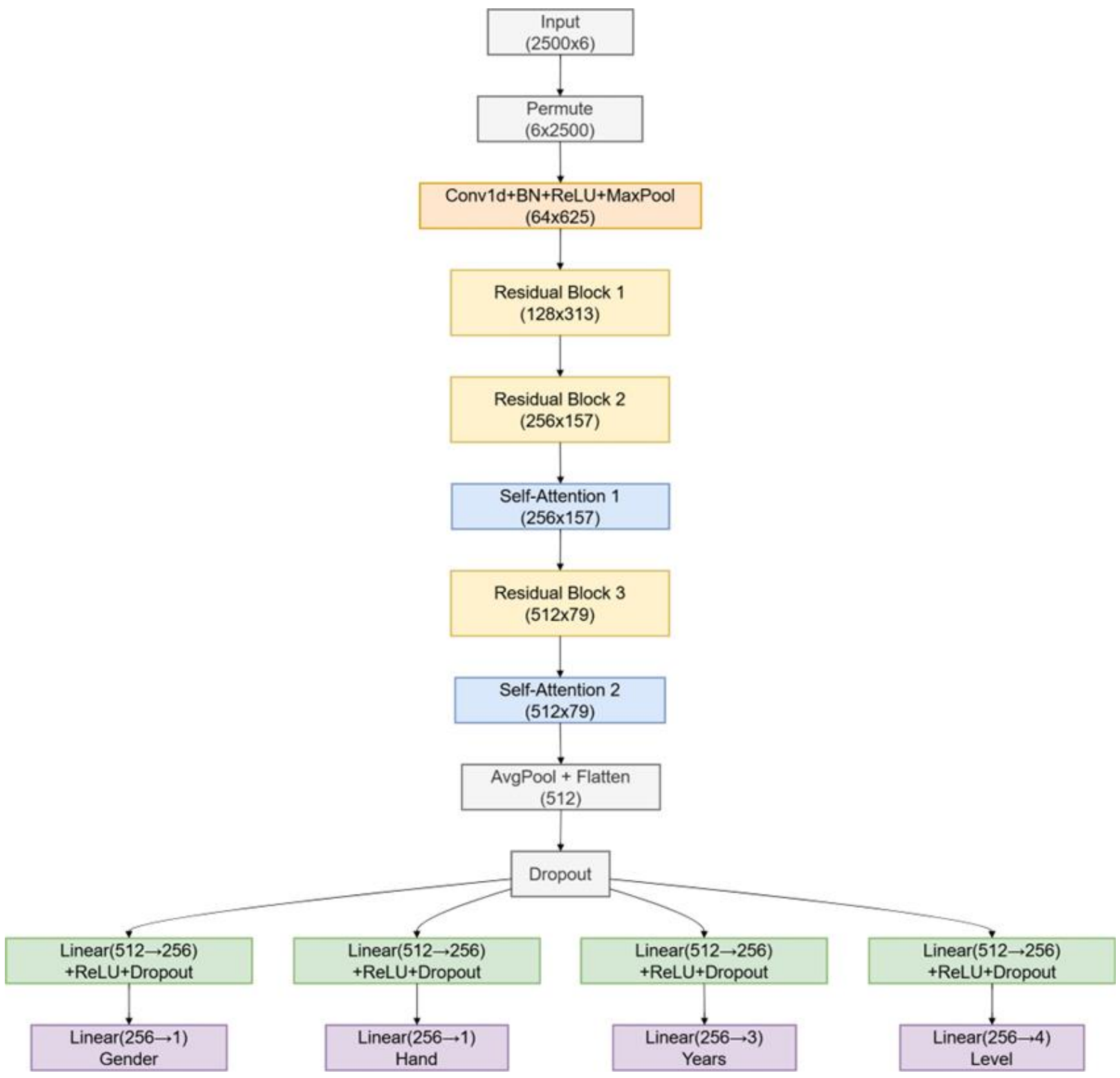

**Fig. 3.** Overview of the proposed TTNet architecture.

**Input and Initial Convolution Layer.** The model's input data is a (16, 2500, 6) tensor of six-axis time series sensor data, where the batch size is 16. The six-axis sensor data corresponds to the acceleration and angular velocity of the x/y/z axes. The input data passes through an initial convolution layer and a max pooling layer for preliminary feature extraction and dimensionality reduction.

**Residual Blocks.** This model consists of three stacked residual blocks. The design of these modules is inspired by ResNet [14]. Each block contains two convolution layers and a shortcut connection, which can effectively address the vanishing gradient problem in deep neural networks. Each residual block gradually increases the number of channels and reduces the sequence length to achieve hierarchical feature extraction. This architecture design is suitable for time-series data from smart table tennis rackets, whose swing movements typically involve multi-stage changes and subtle vibrations. Through the residual blocks, the model can learn both local details and overall patterns, thereby improving its ability to recognize diverse movement patterns.

**Self-Attention Mechanism.** To improve the model's ability to extract global information, we inserted two self-attention layers between the residual blocks, implementing Q/K/V via convolutions and calculating the attention distribution [15]. This enhances the model's capacity to capture long-range sequence features, enabling it to effectively identify correlations between different time points in long sequences. It also helps in identifying key turning points in the swing process, allowing the model to automatically attend to the critical actions or data segments most influential for prediction, such as the beginning and end of the swing.

**Global Pooling and Task Branch Output.** After undergoing multiple transformations, the features are converted into a fixed-length vector through an AdaptiveAvgPool1d to serve as the focus of the entire action. This vector is then fed into four task-specific branches, each consisting of a linear layer, ReLU, Dropout, and a final classification layer to learn the feature mapping for each individual task.

**Multitask Output Design and Loss Computation.** The model output includes four tasks, each branch adopts an independent design, and the appropriate loss function is selected according to the task characteristics, as shown in Table 1. below.

**Table 1.** Multitask output design and corresponding loss function.

| Task | Classification Type | Output Dimension | Loss Function |
|---|---|---|---|
| Gender | Binary Classification | 1 | Focal Loss ($\alpha$=0.25, $\gamma$=2) |
| Playing Hand | Binary Classification | 1 | Focal Loss ($\alpha$=0.25, $\gamma$=2) |
| Years of experience | Multi-class Classification | 3 | Label Smoothing CrossEntropy |
| Player level | Multi-class Classification | 4 | Label Smoothing CrossEntropy |

### 3.4 Novel Method

This study did not rely on pre-trained models but instead designed an architecture with multitask learning capabilities specifically for the six-axis sensor time-series data of smart rackets, demonstrating the following five novel designs:

**Hybrid Architecture Design.** Traditional CNNs are adept at local feature extraction when processing time-series data, but they struggle with long-range dependencies and nonlinear changes. Therefore, we incorporated a self-attention module into the ResNet architecture to enhance the model's ability to capture both local features and global dependencies, thereby effectively processing the high-frequency turns, sudden acceleration changes, and periodic motion patterns commonly found in smart racket data, which helps the model to stably recognize diverse swing styles.

**Multitask Output Branching and Feature Sharing Strategy.** Our model adopts a multitask output design with a shared backbone and task-specific branches. The front end of the model uses a shared feature extraction layer, while the back end consists of four independent task branches. This design improves sample utilization efficiency and facilitates cross-task knowledge transfer, which helps stabilize the performance of tasks with relatively sparse data, such as years of experience and player level classification.

**Loss Function Design.** Due to the imbalance between positive and negative samples in the dataset, different loss functions were applied to each task according to its characteristics:

- Focal Loss: Applied to imbalanced binary classification tasks involving gender and playing hand categories, this method reduces the weight of correctly classified simple samples, allowing the model to focus on learning difficult samples that are hard to distinguish, thereby improving performance on imbalanced datasets.
- Label Smoothing Cross-Entropy: Applied to two multi-category classification tasks, years of experience, and player level, to avoid model overfitting, thereby improving generalization ability and numerical stability [7].

**Two-Stage Training.** We designed a two-stage training process. The first stage focuses on generalization capabilities, introducing a variety of data augmentation methods to improve the model's robustness to action perturbations and sample variability. The second stage involves fine-tuning under reduced interference to enhance the model's convergence and decision boundary accuracy. This strategy achieves a good balance between generalization capabilities and final accuracy.

**Hierarchical Hyperparameter Tuning Strategy.** Following the two-stage training process, we designed a suitable combination of hyperparameters and adjusted them according to the training objectives. The first stage focused on exploration and regularization, while the second stage reduced the various parameters to stabilize convergence.

**Table 2.** Hyperparameter configuration for two-stage training.

| Hyperparameter | Stage 1 | Stage 2 |
|---|---|---|
| Dropout | 0.5 | 0.3 |
| Learning Rate | 1e-4 | 4e-5 |
| Mixup Alpha | 0.3 | 0.05 |
| Label Smoothing | 0.07 | 0.02 |
| Noise Standard Deviation | 0.04 | 0.01 |
| Time Reversal Probability | 0.3 | 0.1 |

## 4 Results

The TTNet architecture achieved stable validation performance and strong generalization capability on the official smart table tennis racket sensor dataset. During training process, we observed that the validation AUC scores approached 1.0, indicating that the model effectively fit the training data features. However, the actual public score reached only approximately 0.84. This result suggests that there may be differences in data distribution between the training set and the testing set, or that the testing set may contain more challenging samples. Therefore, our training strategy shifted from pursuing validation set performance to strengthening the model's generalization ability and robustness, ultimately effectively improving the stability of testing set predictions through data augmentation and staged training approaches. The loss curves from the two-stage training process further demonstrate the effectiveness of our training strategy (see Fig. 4).

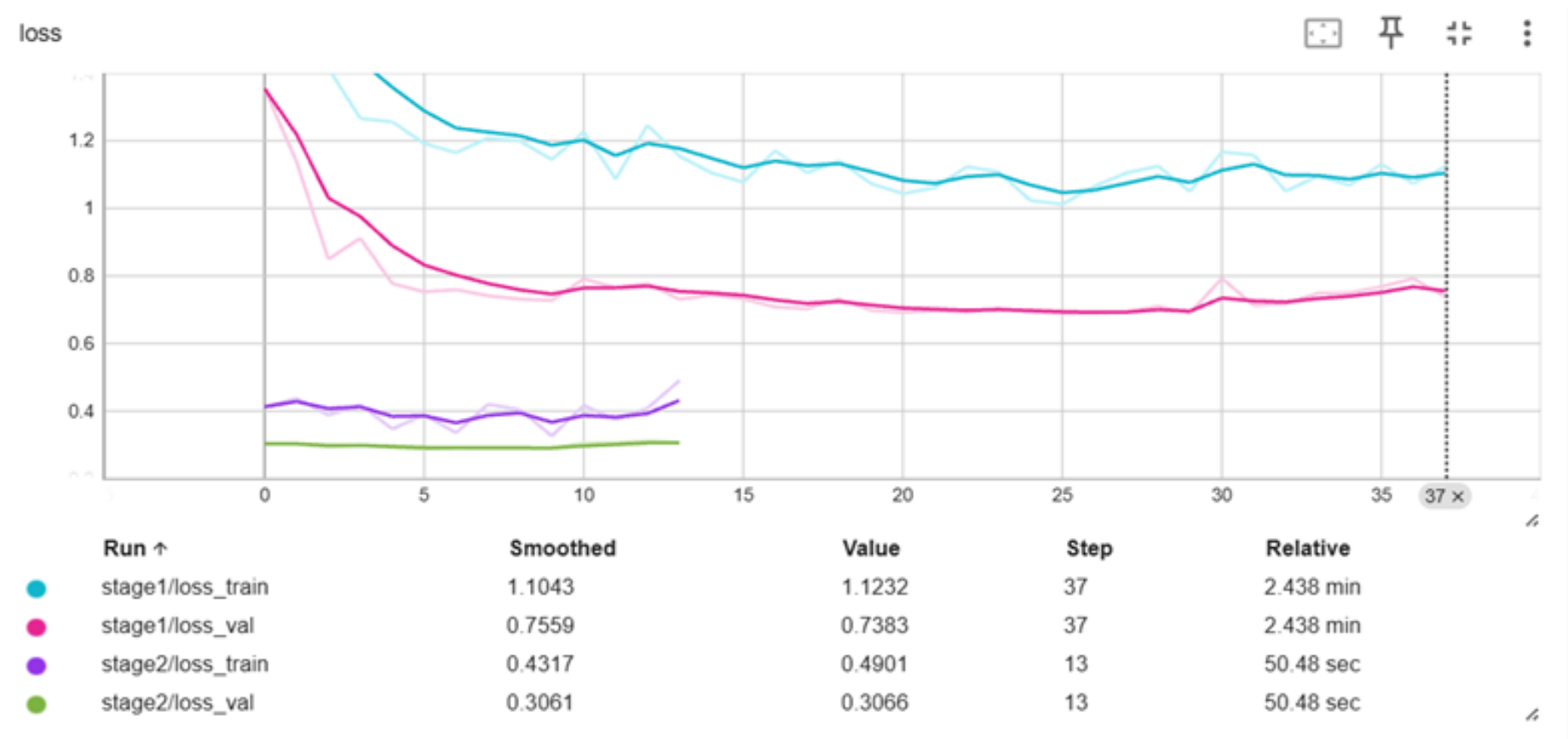

**Fig. 4.** The training loss and validation loss are smoothed to present trends, with the Y-axis representing loss values and the X-axis representing training steps.

# 5 Discussion

## 5.1 Model Evaluation

To validate the overall advantages of the proposed architecture in multitask prediction, we compared TTNet with the Baseline, CatBoost, a standard CNN, and CNN with mode, which incorporates test mode information. The performance of each model on the four prediction tasks is summarized in Table 3.

**Table 3.** Performance comparison of different models on each prediction task.

| Task | Metrix | TTNet (ours) | CNN with mode | CNN | Catboost | Baseline |
|---|---|---|---|---|---|---|
| Gender | ROC AUC | **0.9998** | 0.9977 | 0.9976 | 0.9210 | 0.792 |
| Hand | ROC AUC | **1.0000** | 1.0000 | 1.0000 | 0.9999 | 0.998 |
| Years | OVR ROC AUC | **0.9985** | 0.9962 | 0.9956 | 0.6988 | 0.660 |
| Level | OVR ROC AUC | **0.9997** | 0.9926 | 0.9956 | 0.8493 | 0.822 |

Compared with the baseline model, TTNet shows significant improvements in feature extraction, temporal modeling, and task prediction. The baseline model adopts conventional machine learning methods that rely on manual feature engineering, and it is unable to effectively handle the complex variations in raw time-series data, resulting in relatively low prediction accuracy. CatBoost [16], despite its good training efficiency and generalization ability, cannot directly process high-frequency sequential data. It still relies on engineered features, leading to inferior performance in multitask prediction compared to other models.

The CNN model can directly process raw time-series data and performs well in capturing short-term local features, but it has limited capability in modeling long-range dependencies and global information [17]. CNN with mode learns action distributions under specific conditions by incorporating test mode information, and demonstrates a slight improvement in prediction accuracy. However, it remains a single-task architecture and cannot integrate cross-task features.

In contrast, TTNet integrates residual structures and self-attention modules, providing both local and global feature modeling capabilities. Through multitask shared representation, it effectively integrates data across tasks, enhancing generalization and prediction stability, particularly demonstrating a clear advantage over other methods in multi-class prediction tasks.

## 5.2 Ablation Study

To evaluate the actual contribution of each key method in the Full Model to the final performance and quantify the effectiveness of each key method. We established the complete model containing all designs as E0 as the baseline and created a series of experimental groups from E1 to E6. Each experimental group removes or modifies a single key method from the Full Model, with specific designs shown in Table 4.

**Table 4.** Ablation study experimental design.

| Experiment ID | Experiment Name | Removed/Modified Component | Purpose |
|---|---|---|---|
| E0 | Full Model | None | Establish baseline |
| E1 | No Self-Attention | Remove self-attention module | Verify contribution of capturing temporal long-range dependencies |
| E2 | No Residual Connections | Remove residual connections | Verify effectiveness of stabilizing deep network training |
| E3 | No Mixup Augmentation | Disable Mixup | Evaluate the contribution of Mixup to generalization |
| E4 | No Time Reversal | Disable time-reversal augmentation | Evaluate the effectiveness of domain-specific augmentation |
| E5 | No Focal Loss | Replace with standard BCE Loss | Verify advantages in handling class imbalance |
| E6 | Single-Stage Training | Remove the two-stage training strategy | Evaluate importance of staged training in generalization stability |

All experiments were conducted under the same environment with fixed random seeds to ensure fair comparison. We adopted validation loss and ROC AUC scores for binary classification tasks and OVR ROC AUC scores for multi-class classification tasks as core evaluation metrics.

### 5.3 Results and Analysis

According to Table 5, the ablation studies significantly affected model performance. The most obvious difference in experimental results is E6, whose validation loss increased to 0.27351, deteriorating by 9.2% compared to the Full Model. This reflects the most severe performance degradation among all experiments, especially in terms of a significant decline in the AUC metric for the Level task. These findings highlight the importance of our proposed two-stage training strategy, with the first stage stronger data augmentation and the second stage fine-tuning optimization. This strategy plays a key role in enhancing the model's stability and generalization ability.

In terms of core architecture, E1 and E2 resulted in 5.8% and 3.1% loss increases, respectively. These results validate the importance of self-attention for capturing long-range dependencies and the contribution of residual structures in preventing gradient vanishing and strengthening deep representation capabilities.

E5 achieved good AUC scores on the gender task but showed a significant overall loss increase, demonstrating that Focal Loss provides substantial benefits in handling sample imbalance and difficult samples.

E3 resulted in the lowest loss but caused a significant AUC decline in the gender task. Reflects that the Mixup method effectively suppressed model overfitting on specific tasks, thereby ensuring overall model robustness and generalization ability.

E4 had the least impact on overall performance, though it led to a slight increase in validation loss. Confirms that this domain-specific augmentation method still provides a beneficial contribution.

**Table 5.** Performance comparison in ablation study.

| Experiment ID | Experiment Name | Validation Loss | AUC (Gender) | AUC (Hand) | AUC (Years) | AUC (Level) |
|---|---|---|---|---|---|---|
| E0 | Full Mode | 0.25042 | 0.99963 | **1.00000** | **1.00000** | **1.00000** |
| E1 | No Self-Attention | 0.26496 | 0.99953 | 1.00000 | 0.99990 | 0.99996 |
| E2 | No Residual Connections | 0.25826 | 0.99977 | 1.00000 | 0.99999 | 0.99998 |
| E3 | No Mixup Augmentation | **0.24783** | 0.99865 | 1.00000 | 1.00000 | 1.00000 |
| E4 | No Time Reversal | 0.25132 | 0.99963 | 1.00000 | 1.00000 | 0.99993 |
| E5 | No Focal Loss | 0.25998 | **1.00000** | 1.00000 | 0.99999 | 0.99999 |
| E6 | Single-Stage Training | 0.27351 | 0.99911 | 1.00000 | 1.00000 | 0.99522 |

## 6 Conclusion

This study introduces a deep learning model with multitask learning capability, TTNet, which can simultaneously predict gender, playing hand, years of experience, and player level from smart table tennis racket data. The model integrates a two-stage training strategy with targeted data augmentation techniques, demonstrates strong stability and generalization when faced with class imbalance and motion variability. The model ultimately achieved an impressive second-place result in the AI CUP 2025 competition, validating the practical potential of the proposed approach in sports sensor data analysis and smart training system applications.